\documentclass[runningheads]{llncs}

\usepackage[mobile]{eccv}

\usepackage{eccvabbrv}

\usepackage{graphicx}
\usepackage{booktabs}

\usepackage{multirow}
\usepackage{threeparttable}
\usepackage{xfrac}
\usepackage[table]{xcolor}
\usepackage{twemojis}

\usepackage[accsupp]{axessibility}  % Improves PDF readability for those with disabilities.

\usepackage[pagebackref,breaklinks,colorlinks,citecolor=eccvblue]{hyperref}
\usepackage{orcidlink}

\begin{document}

\title{Word-Anchored Temporal Forgery Localization} 


\author{
Tianyi Wang\inst{1},
Xi Shao\inst{2},
Harry Cheng\inst{1},
Yinglong Wang\inst{3}, \\
\and Mohan Kankanhalli\inst{1}
}

\authorrunning{T.~Wang et al.}

\institute{National University of Singapore \and
Nanjing University of Posts and Telecommunications \and
Qilu University of Technology (Shandong Academy of Sciences)
% \url{http://www.springer.com/gp/computer-science/lncs}
}

\maketitle

\begin{abstract}
  Current temporal forgery localization (TFL) approaches typically rely on temporal boundary regression or continuous frame-level anomaly detection paradigms to derive candidate forgery proposals. However, they suffer not only from feature granularity misalignment but also from costly computation. To address these issues, we propose word-anchored temporal forgery localization (WAFL), a novel paradigm that shifts the TFL task from temporal regression and continuous localization to discrete word-level binary classification. Specifically, we first analyze the essence of temporal forgeries and identify the minimum meaningful forgery units, word tokens, and then align data preprocessing with the natural linguistic boundaries of speech. To adapt powerful pre-trained foundation backbones for feature extraction, we introduce the forensic feature realignment (FFR) module, mapping representations from the pre-trained semantic space to a discriminative forensic manifold. This allows subsequent lightweight linear classifiers to efficiently perform binary classification and accomplish the TFL task. Furthermore, to overcome the extreme class imbalance inherent to forgery detection, we design the artifact-centric asymmetric (ACA) loss, which breaks the standard precision-recall trade-off by dynamically suppressing overwhelming authentic gradients while asymmetrically prioritizing subtle forensic artifacts. Extensive experiments demonstrate that WAFL significantly outperforms state-of-the-art approaches in localization performance under both in- and cross-dataset settings, while requiring substantially fewer learnable parameters and operating at high computational efficiency. 
\end{abstract}

\section{Introduction}
\label{sec:introduction}

The ongoing battle between Deepfake forgeries and forensic countermeasures has yielded promising solutions from passive~\cite{NoiseDF,ForensicAdapter,luo2023beyond,Effort} and proactive~\cite{IDP-Mark,FractalForensics,SVS-WM} perspectives for Deepfake detection on images and full videos~\cite{Deepfake2024Wang_ComputingSurveys}. However, these global classifiers are generally unequipped to distinguish audio-visual temporal forgeries, where Deepfake manipulates only segments of a video.

In real-life scenarios, malicious actors usually favor temporal forgeries over full-video manipulation to maximize deceptive realism while minimizing computational overhead. Consequently, recent temporal forgery localization (TFL) approaches aim to identify precise boundaries of the manipulated segments that are seamlessly wrapped within the natural temporal flow. To achieve this, existing paradigms mainly adapt temporal action localization (TAL) networks for boundary regression~\cite{BA-TFD,BA-TFD+,UMMAFormer,Pindrop,Localizing2025Chen}, or employ continuous frame-level anomaly detection and cross-modal analyses to identify temporal inconsistencies~\cite{Context2025Yin,AudioVisual2024Liu,AuViRe,RegQAV,Intra-modal2025Anshul}. In a nutshell, they typically extract sequence-level features from pre-trained backbones optimized for semantic recognition and utilize continuous sliding windows or dense boundary matching networks to derive candidate forgery proposals. However, attempting to localize discrete high-frequency forensic artifacts using frozen semantic features and TAL frameworks, which are designed for continuous low-frequency semantic-level feature analysis, introduces performance bottlenecks due to feature granularity misalignment. Furthermore, dedicating excessive computational power to evaluating continuous frames or regressing precise boundaries inevitably limits the focus on distinguishing forensic artifacts. 

\begin{figure}[tb]
  \centering
  \includegraphics[width=\textwidth]{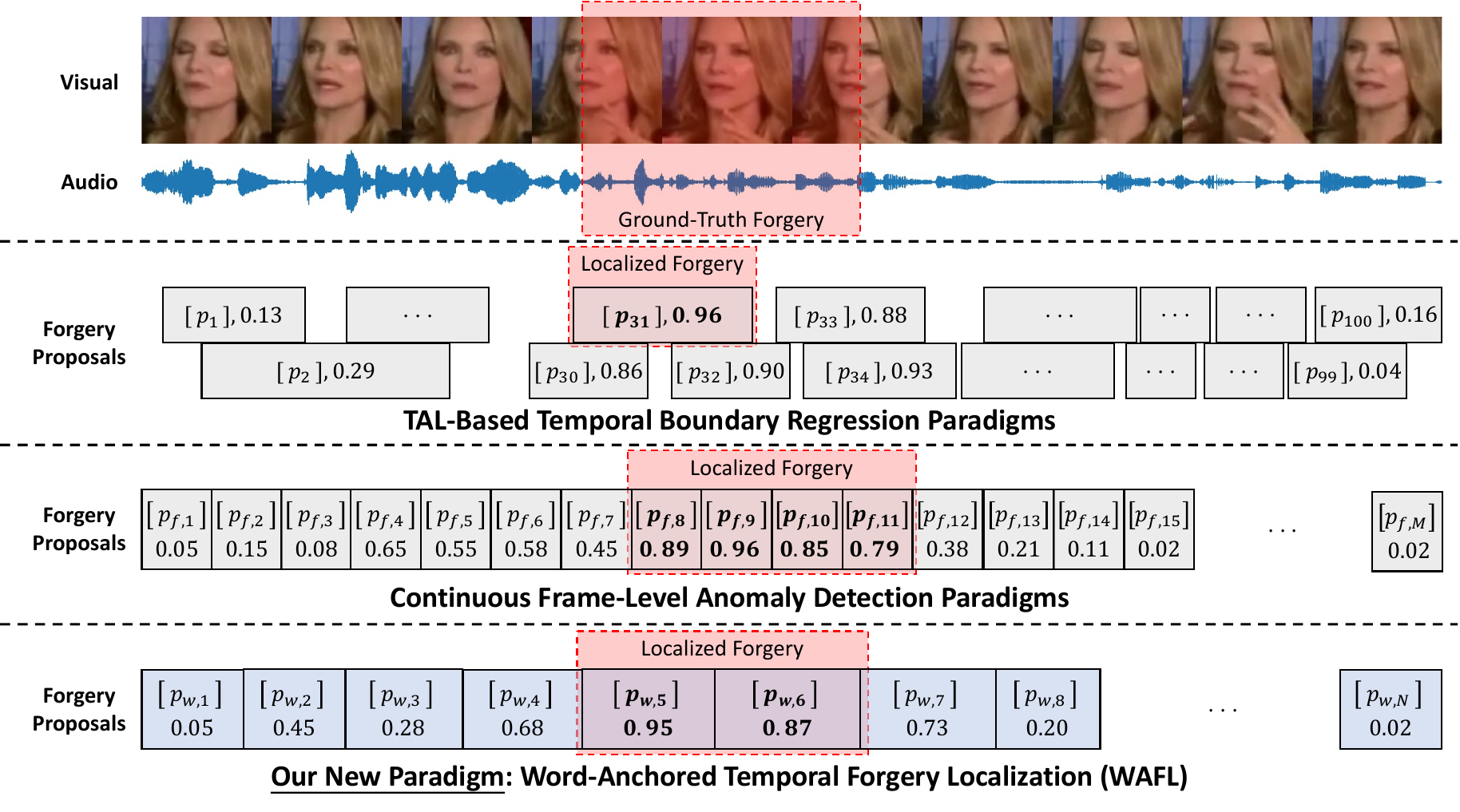}
  \caption{Demonstration of the proposed new paradigm compared to current state-of-the-art approaches. }%State-of-the-art paradigms perform continuous localization and regression to finalize the overlapped forgery proposals, whereas our novel paradigm generates discrete word-token-based proposals.
  \label{fig:motivation}
\end{figure}

In this paper, we analyze the essence of temporal Deepfake forgeries and propose a novel word-anchored paradigm for the TFL task. Considering that the ultimate objective of audio-visual temporal Deepfake manipulations is to deceive the audience by altering semantic meaning, they are inherently bound to the natural rhythm of human speech. Therefore, we argue that the discrete word tokens are the minimum meaningful units of temporal forgeries in videos. For instance, when a malicious actor aims to modify the information conveyed in a video, the Deepfake manipulation commonly occurs at the lexical level. In other words, tampering with arbitrary continuous frames that are only a fraction of a spoken word brings trivial outcomes regarding semantic deception and leaves obvious forensic discontinuities that can be easily observed. 

Building upon this insight, we introduce word-anchored temporal forgery localization, WAFL in short, a novel paradigm that explicitly eliminates dense, overlapping proposals of existing frameworks, transforming the continuous localization task into a discrete binary classification task on word tokens (\cref{fig:motivation}). Specifically, we first adopt the off-the-shelf speech-to-text tool to segment word tokens within each video accordingly. Then, we establish a forensic feature realignment (FFR) module consisting of two encoders for feature extraction, one for each modality. Unlike existing approaches, upon leveraging the frozen foundation models for visual and audio modalities, we explicitly map representations from the pre-trained low-frequency semantic space to a highly discriminative forensic manifold. This allows a subsequent group of significantly lightweight linear heads to efficiently perform the final binary classification on each word token, and the forgery proposals are then derived alongside. During training, to address the extreme imbalance where authentic word tokens vastly outnumber Deepfake ones, we propose the artifact-centric asymmetric (ACA) loss that breaks the conventional precision-recall trade-off by enforcing a strict penalty for fake samples while dynamically suppressing the overwhelming gradients of easy real samples. Extensive experiments on the LAV-DF~\cite{BA-TFD} and AV-Deepfake1M~\cite{AV-Deepfake1M} datasets demonstrate the superiority of the proposed WAFL paradigm for in-dataset and cross-dataset settings, while requiring substantially fewer learnable parameters. The contributions of this work are as follows:
\begin{itemize}
\item We analyze the essence of temporal forgeries and propose a novel word-anchored paradigm that shifts the TFL task from boundary regression and continuous localization to precise, discrete binary classification.
\item We introduce a forensic feature realignment module that projects pre-trained semantic space onto a discriminative high-frequency forensic manifold, which then allows lightweight linear heads to efficiently derive promising outcomes.
\item We design the artifact-centric asymmetric loss to dynamically mitigate the extreme class imbalance issue, leading to state-of-the-art localization performance on various benchmark datasets.
\end{itemize}

\section{Related Work}
\label{sec:related_work}

Ever since the first occurrence of the temporal forgery localization (TFL) task, cutting-edge approaches have experienced a diverse evolution. Early methods attempt to address the TFL task analogously to temporal action localization (TAL), mainly relying on continuous visual streams for manipulation boundary regression. Specifically, Cai \etal~\cite{BA-TFD,BA-TFD+} leveraged a multiscale vision transformer to extract continuous features, followed by direct porting to the boundary matching network~\cite{BMN}. UMMAFormer~\cite{UMMAFormer} evolves the TAL paradigm via parallel cross-attention and feature pyramid networks by sliding multiscale continuous windows over the video. Similarly, Chen \etal~\cite{Localizing2025Chen} proposed HBMNet, which utilizes a coarse-to-fine refinement strategy to produce boundary-matching maps and refine proposal boundaries through bidirectional probability modeling. Klein \etal~\cite{Pindrop} further adopted a stronger TAL backbone, ActionFormer~\cite{ActionFormer}, along with advanced transformers and multi-stage temporal convolutional networks to output frame-wise continuous boundary proposals. While relying on powerful backend modules, they are trapped by the granularity misalignment between semantic and forensic manifolds. 

Subsequent studies are shifted toward continuous frame-level anomaly detection. Yin \etal~\cite{Context2025Yin} formulated TFL via contrastive learning and designed a context-aware perception layer to justify the distance between an instant feature and the global video context. Similarly, Liu \etal~\cite{AudioVisual2024Liu} employed self-attention~\cite{Transformer} for embedding-level fusion, multi-dimensional contrastive loss to enforce sequence consistency, and 1D convolutions to output frame-level anomaly scores. As Deepfake manipulation techniques evolve, the research domain has advanced toward cross-modal interactions to look for inconsistencies between visual and audio modalities. For example, a query-based learning framework with learnable registers within the attention mechanism, RegQAV~\cite{RegQAV}, forces fine-grained alignment between heterogeneous audio and visual signals. DiMoDif~\cite{DiMoDif} analyzes cross-modal discourse differentiation to localize mismatches between modalities at the semantic level. AuViRe~\cite{AuViRe} proposes generative cross-modal reconstruction between visual and audio modalities. Forgeries are localized when reconstruction errors occur. Recently, Anshul \etal~\cite{Intra-modal2025Anshul} introduced a two-stage pipeline that leverages intra-modal and cross-modal synchronization pre-trained on real data to detect temporal forgeries as disruptions in natural signal continuity. Alternatively, AVH-Align~\cite{AVH-Align} and MDP~\cite{MDP} both adopt the semi-supervised training pipeline to drop boundary regression, where the former relies on temporal synchronization between lip movements and the audio stream, while the latter introduces temporal property-preserving cross-modal attention and gathers frame-level predictions for the video-level prediction without timestamp annotations participating. Although sophisticated fusion techniques have brought advancement in localization performance, they process forgery artifacts as continuous temporal streams or consecutive visual frames, still ignoring the natural linguistic rhythm of speech, making these paradigms suffer severe computational overhead in finalizing precise boundaries rather than distinguishing real and fake. To overcome the research gap, in this paper, we analyze the essence of temporal forgeries and propose a new word-anchored paradigm with respect to the linguistic boundaries of human speech. 

\section{Methodology}
\label{sec:methodology}

\subsection{Overview}
\label{sec:overview}

\begin{figure}[tb]
    \centering
    \includegraphics[width=\textwidth]{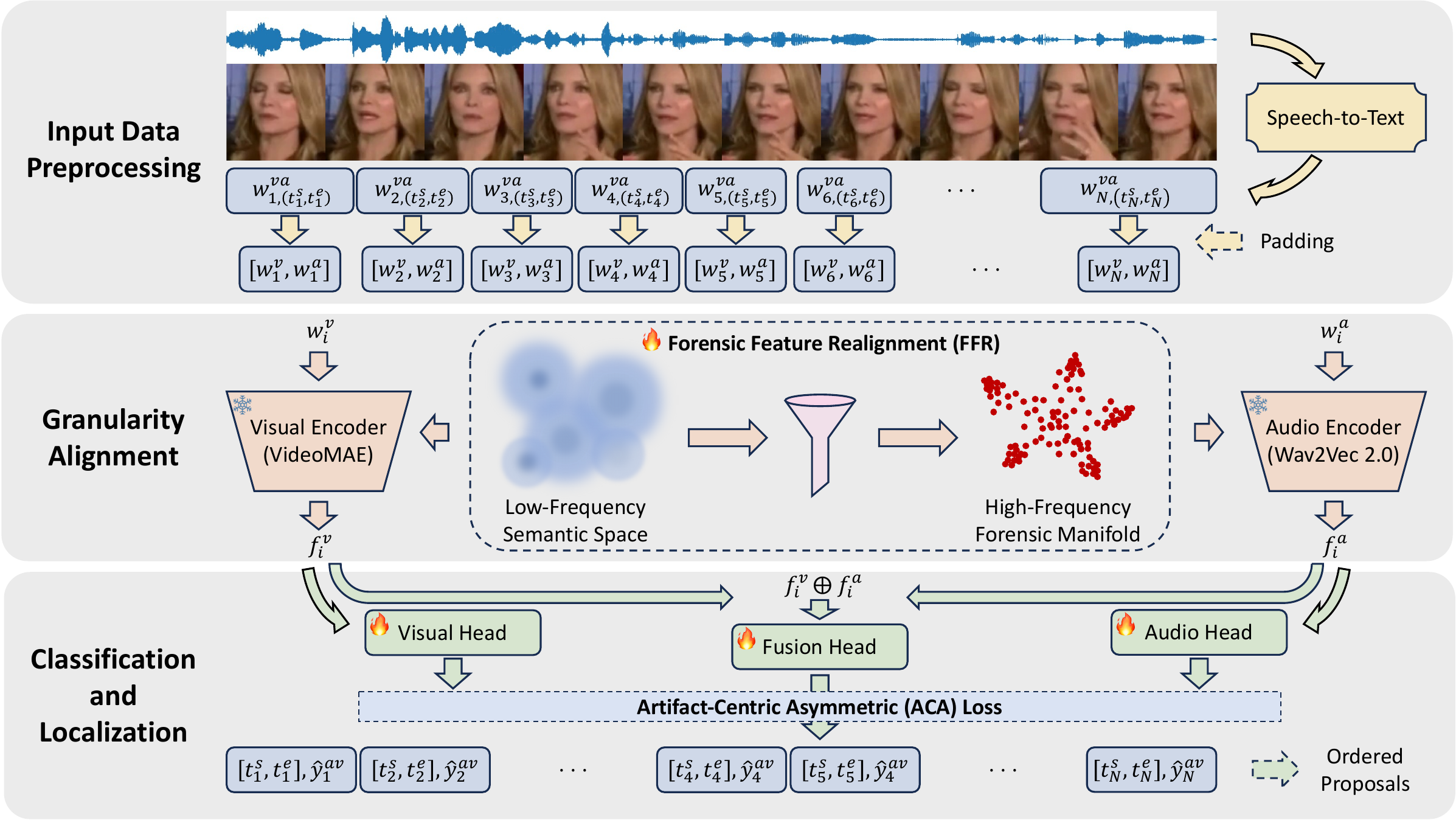}
    \caption{Workflow of the proposed framework for temporal forgery localization. }
    \label{fig:workflow}
\end{figure}

The proposed word-anchored temporal forgery localization (WAFL) paradigm shifts the temporal forgery localization (TFL) task from boundary regression and continuous localization to discrete binary classification on word tokens. As illustrated in \cref{fig:workflow}, given an input audio-visual video, WAFL operates in three sequential stages, namely data preprocessing, forensic feature realignment (\cref{sec:forensic_feature_realignment}), and temporal forgery proposal generation (\cref{sec:temporal_forgery_proposal_generation}). To obtain the discrete word tokens, we first apply an off-the-shelf speech-to-text alignment tool\footnote{Google Speech-to-Text.} to the audio track, deriving the transcript alongside the start and end timestamps for each recognized word. Then, for a video with $N$ word tokens, we perform segmentation on the video accordingly, denoted as $W=\{[w^v_i, w^a_i]\}_{i=1}^N$, where each word token is represented by a pair of visual and audio segments, $w^v_i$ and $w^a_i$, respectively. This operation discretizes the video into non-overlapping lexical units, avoiding the high computational cost of dense sliding windows. Lastly, since the natural duration varies across words, the raw visual and audio segments are padded following reflection and trailing strategies to a pre-defined fixed temporal length. 

The uniformly formatted tokens are then fed into the forensic feature realignment (FFR) module, where the visual and audio encoders extract forensic artifact features from $w_i^v$ and $w_i^a$, respectively, for $1 \le i \le N$. Thereafter, the features are passed through three classifier heads, designated for visual, audio, and visual-audio fused features, respectively, where all three participate in training, and only the fusion head is used for classification on word tokens. Finally, the temporal forgery proposals for each video are ordered by classification scores and then used to evaluate the performance on the AP@IoU and AR@N metrics. Additionally, during training, the model is optimized via the artifact-centric asymmetric (ACA) loss, specifically designed to mitigate the extreme class imbalance issue between real and fake word tokens.

\subsection{Forensic Feature Realignment}
\label{sec:forensic_feature_realignment}

To extract robust multimodal representations, rather than training from scratch, we employ two high-capacity foundation models as feature extraction backbones, VideoMAE~\cite{VideoMAE} for the visual modality and Wav2Vec 2.0~\cite{Wav2Vec_V2} for the audio modality. This allows us to leverage their considerable capacity for capturing intricate spatio-temporal dynamics and acoustic nuances, which we subsequently specialize in forensic artifacts via our proposed forensic feature realignment module. For a given word token, the visual segment, $w_i^v$, is formatted as an $M$-frame clip with a spatial resolution of $224 \times 224$, and the audio segment, $w_i^a$, is processed as a 1D waveform. While these foundation models provide rich feature spaces, they are designed, pre-trained, and optimized for semantic tasks, such as action recognition and speech transcription, respectively. Therefore, their frozen feature spaces exhibit inconsistencies when directly transferred to subtle high-frequency artifacts of temporal forgeries. Conversely, an end-to-end fine-tuning is computationally heavy and prone to overfitting.

To resolve this granularity misalignment, we introduce the forensic feature realignment (FFR) module. Instead of updating the entire network, we project representations from the pre-trained semantic space onto a highly discriminative forensic manifold by repurposing Low-Rank Adaptation (LoRA)~\cite{LoRA}, beyond simple adaptation. In particular, we freeze all original parameters of the pre-trained backbones and inject trainable rank-decomposition matrices into query and value projection layers of the transformer blocks. Moreover, to prevent granularity alignment from overfitting to dataset-specific noise, we present stochastic artifact regularization. Specifically, the forward pass with frozen weight, $W_0\in\mathbb{R}^{d\times k}$, is reformulated as
\begin{equation}
h = W_0x + \frac{\alpha}{r} \Phi^\uparrow \Phi^\downarrow \delta(x),
\label{eq:lora}
\end{equation}
where $\Phi^\uparrow \in \mathbb{R}^{d \times r}$ and $\Phi^\downarrow \in \mathbb{R}^{r \times k}$ are learnable realignment matrices that project the input feature $x$ into a low-rank bottleneck $r \ll \min(d,k)$, with the constant scaling factor $\frac{\alpha}{r}$ for intensity control. Notably, $\delta(\cdot)$ applies stochastic dropout to $x$, forcing $\Phi^\uparrow$ and $\Phi^\downarrow$ to learn generalized robust manipulation cues rather than memorizing local noise patterns. Finally, the outputs are applied with global average pooling to match the dimension of the final classification. 

\subsection{Temporal Forgery Proposal Generation}
\label{sec:temporal_forgery_proposal_generation}

After FFR, each word token is represented as a pair of discriminative forensic feature vectors, $f_i^v \in \mathbb{R}^{d}$ and $f_i^a \in \mathbb{R}^{d}$, where $d$ is the shared projection dimension. Meanwhile, the aggregated multimodal forensic feature $f_i^{va}$ is derived via concatenation, $f_i^v \oplus f_i^a$. We then build three independent lightweight linear classifiers, namely the visual head $h^v(\cdot)$, audio head $h^a(\cdot)$, and fusion head $h^{va}(\cdot)$, operating on features $f_i^v$, $f_i^a$, and $f_i^{va}$, respectively, to compute the forgery prediction scores following
\begin{equation}
\hat{y}_i^m = \sigma(W^m f_i^m +b^m) \text{ for } m \in \{ a, v, av \},
\label{eq:heads}
\end{equation}
where $W(\cdot)$ and $b$ are parameters of each linear projection, and $\sigma(\cdot)$ represents the sigmoid activation to derive the forgery prediction scores. 

During training, all three heads are supervised simultaneously. In this way, each encoder in FFR independently mines robust modality-specific forensic artifacts, preventing the multimodal representation from degenerating or over-relying on a single dominant modality. Conversely, once well-trained, only the fused forgery score, $\hat{y}_i^{av}\in[0,1]$, is adopted for inference. Thanks to the novel WAFL paradigm that operates on discrete lexical units rather than continuous video streams, we generate the final temporal forgery proposals regarding the timestamps of word tokens and formulate the tuples $P=\{([t_i^s, t_i^e], \hat{y}_i^{av})\}_{i=1}^N$ for a video containing $N$ word tokens each with start time $t_i^s$ and end time $t_i^e$, along with the proposal score $\hat{y}_i^{av}$. Subsequently, the proposals are ranked by the prediction scores to compute AP@IoU and AR@N metrics for evaluation. 

\subsection{Artifact-Centric Asymmetric Loss}

Following the proposed WAFL paradigm, the nature of temporal forgeries introduces extreme class imbalance at the word level, leading to a potential explosion of false positives during training. Specifically, in a partially manipulated Deepfake video, most word tokens remain authentic, whereas only a few words are forged. Although a pre-defined balanced data sampler can ensure fairness in the overall distribution, the occurrence frequencies differ substantially between real and fake samples. In the TFL task, we expect the correct proposals to rank the highest by confidence scores. Therefore, to break the precision-recall trade-off, we propose the artifact-centric asymmetric (ACA) loss as 
\begin{equation}
\mathcal{L}_{ACA} = 
\begin{cases}
-(1-p)^{\gamma_+}\log{(p)} & \text{if } y=1, \\
-p_m^{\gamma_-}\log{(1-p_m)} & \text{if } y=0. 
\end{cases}
\label{eq:aca_loss}
\end{equation}
Specifically, $\gamma_+$ and $\gamma_-$ are asymmetric modulation factors for the fake and real classes, respectively. In the extreme imbalance condition with real samples dominating, any reduction in the gradient penalty risks missing subtle forensic artifacts, severely affecting the recall. On the other hand, the model is expected to smoothly ignore the massive, harmless variation in natural contents to avoid false positives. Therefore, we suppress the gradient of the real samples by defining $\gamma_- \gg \gamma_+$. Furthermore, to avoid wasting optimization capacity on the easy real samples that are already well-learned, we introduce a probability margin, $\mu$, and compute the margin-shifter probability, $p_m$, for the negative class following
\begin{equation}
p_m=\max{(p-\mu,0)}.
\label{eq:margin_shift}
\end{equation}
When a real sample is correctly predicted with a probability smaller than the threshold, $\mu$, its contribution to the loss gradient is explicitly zeroed out. 

In this work, to handle cases where only one modality contains fake segments and to avoid any modality dominating the entire model, our WAFL framework comprises three distinct classifier heads. During training, the ACA loss is computed independently for each head, forcing both the unimodal encoders and the fused representations to favorably isolate forensic artifacts. The overall loss is denoted as
\begin{equation}
\mathcal{L}=\mathcal{L}_{ACA}^v+\mathcal{L}_{ACA}^a+\mathcal{L}_{ACA}^{va}.
\end{equation}
As a result, the ACA loss strictly prioritizes rare, high-frequency forensic anomalies while dynamically filtering out the overwhelming noise of easy authentic samples.

\section{Experiments}
\label{sec:experiments}

\subsection{Implementation Details}
\label{sec:implementation_details}

\noindent\textbf{Datasets.} In experiments, we adopted the popular benchmark datasets, LAV-DF~\cite{BA-TFD} and AV-Deepfake1M~\cite{AV-Deepfake1M}, for temporal forgery localization. They both provide official data splits, where LAV-DF contains 78,703, 31,350, and 26,100 samples, and AV-Deepfake1M contains 746,180, 57,340, and 343,240 samples, respectively, for training, validation, and testing. In this paper, we leveraged the testing split of LAV-DF and the validation split of AV-Deepfake1M\footnote{Labels for the AV-Deepfake1M testing split are not publicly available. } for performance evaluation. 

\noindent\textbf{Hyperparameters.} To align with the pre-trained visual encoder, VideoMAE~\cite{VideoMAE}, all visual and audio segments are padded to 0.64 seconds, i.e., 16 visual frames and 10,240 audio samples. For the FFR module, we set $r=8$ and $\alpha=16$ in \cref{eq:lora}. $\Phi^{\uparrow}$ is initialized to zero to strictly match the pre-trained semantic output, while $\Phi^{\downarrow}$ is initialized using Kaiming uniform distribution~\cite{KaimingInit} to preserve capacity for diverse forensic feature learning. For the ACA loss, we set $\mu=5e-2$ with asymmetric modulation factors $\gamma_+=0$ and $\gamma_-=4$, to dynamically suppress gradient noise from the easy real samples. Although unable to balance occurrence frequencies, we implemented a custom balanced sampler to ensure a $1:1$ ratio of real to fake tokens within each training iteration. During training, an AdamW optimizer~\cite{AdamW} is adopted with a learning rate of $8e-4$ and a weight decay of $1e-4$. We trained the model for 25k iterations with the first 2.5k to linearly warm up the device, setting a batch size of 64 on 8 NVIDIA A100 GPUs.

\noindent\textbf{Evaluation Metrics.} Following the convention, we evaluated the TFL performance using average precision (AP@IoU) and average recall (AR@N), where IoU$=\{0.5, 0.75, 0.95\}$ and N$=\{100, 50, 20, 10, 5, 2\}$\footnote{For LAV-DF we compute AR@2 and for AV-Deepfake1M we do AR@5 to match their most forgery segments per video.}, respectively. 

\subsection{Temporal Forgery Localization Performance}
\label{sec:in-dataset_performance}

To evaluate the effectiveness of the proposed WAFL paradigm, we first conducted an in-dataset comparison, i.e., trained and evaluated on the same dataset, against current state-of-the-art TFL approaches on the LAV-DF and AV-Deepfake1M datasets. Results are quantitatively presented in \cref{tab:lavdf} and \cref{tab:avdeepfake1m}, respectively. In general, WAFL successfully achieves superior performance across all metrics. Specifically, regarding AP@IoU, WAFL delivers highly competitive performance, achieving the best AP@0.5 values of 99.76\% on LAV-DF and 97.26\% on AV-Deepfake1M, despite the state-of-the-art competitors already being outstandingly promising. A closer look at the performance of earlier paradigms like BA-TFD~\cite{BA-TFD} and BA-TFD+~\cite{BA-TFD+} reveals the instability of traditional regression-based localization paradigms. While performing with reasonable statistics on LAV-DF, when facing AV-Deepfake1M, these models struggle to surpass 45\% even for the tolerant AP@0.5 metric, highlighting that basic temporal regression may be insufficient for large-scale, complex Deepfake scenarios.

\begin{table}[t!]
\caption{Temporal forgery localization performance (\%) on LAV-DF under in-dataset setting. \textbf{Best} and \underline{second-best} performance marked in \textbf{bold} and \underline{underline}. }
\label{tab:lavdf}
\centering
\resizebox{\textwidth}{!}{
\begin{tabular}{@{}lcccccccc@{}}
\toprule
 & AP@0.5 & AP@0.75 & AP@0.95 & AR@100 & AR@50 & AR@20 & AR@10 & AR@2 \\
\midrule
BA-TFD~\cite{BA-TFD} & 79.15 & 38.57 & 0.24 & 66.90 & 64.08 & 60.77& 58.42 & 53.33 \\
BA-TFD+~\cite{BA-TFD+} & 96.30 & 84.96 & 4.44 & 81.62 & 80.48 & 79.40 & 78.75 & 75.85 \\
UMMAFormer~\cite{UMMAFormer} & \underline{98.83} & \underline{95.53} & 37.73 & 92.49 & 92.49 & 92.43 & 92.11 & 87.14 \\
MDP~\cite{MDP} & 53.36 & 37.12 & 0.02 & 40.92 & 40.92 & 40.53 & 38.45 & 25.69 \\
DiMoDif~\cite{DiMoDif} & 95.50 & 87.91 & 20.61 & 94.17 & 93.71 & 92.73 & 91.35 & 83.12 \\
AuViRe~\cite{AuViRe} & 98.16 & 94.74 & \underline{45.83} & \underline{94.42} & \underline{94.04} & \underline{93.50} & \underline{92.79} & \underline{87.73} \\
\rowcolor{gray!20} WAFL (ours) & \textbf{99.76} & \textbf{99.31} & \textbf{99.31} & \textbf{99.74} & \textbf{99.74} & \textbf{99.74} & \textbf{99.74} & \textbf{99.73} \\
\bottomrule
\end{tabular}
}
\end{table}

\begin{table}[t!]
\caption{Temporal forgery localization performance (\%) on AV-Deepfake1M under in-dataset setting. \textbf{Best} and \underline{second-best} performance marked in \textbf{bold} and \underline{underline}. }
\begin{threeparttable}
\label{tab:avdeepfake1m}
\centering
\resizebox{\textwidth}{!}{
\begin{tabular}{@{}lcccccccc@{}}
\toprule
 & AP@0.5 & AP@0.75 & AP@0.95 & AR@100 & AR@50 & AR@20 & AR@10 & AR@5 \\
\midrule
BA-TFD$^\dagger$~\cite{BA-TFD} & 37.37 & 6.34 & 0.02 & -- & 45.55 & 34.95 & 30.66 & 26.82 \\
BA-TFD+$^\dagger$~\cite{BA-TFD+} & 44.42 & 13.64 & 0.03 & -- & 48.86 & 40.37 & 34.67 & 29.88 \\
UMMAFormer$^\dagger$~\cite{UMMAFormer} & 51.64 & 28.07 & 1.58 & -- & 44.07 & 43.45 & 42.09 & 40.27 \\
MDP~\cite{MDP} & 0.01 & 0.00 & 0.00 & 0.03 & 0.03 & 0.03 & 0.02 & 0.02 \\
DiMoDif~\cite{DiMoDif} & 96.04 & 89.23 & 6.85 & \underline{89.24} & \underline{88.69} & \underline{87.68} & \underline{86.39} & 84.32 \\
AuViRe~\cite{AuViRe} & \underline{96.96} & \underline{90.04} & \underline{12.64} & 86.66 & 86.51 & 86.18 & 85.67 & \underline{84.72} \\
\rowcolor{gray!20} WAFL (ours) & \textbf{97.26} & \textbf{97.25} & \textbf{97.24} & \textbf{99.99} & \textbf{99.99} & \textbf{99.98} & \textbf{99.90} & \textbf{99.69} \\
\bottomrule
\end{tabular}
}
\begin{tablenotes}
\item $^\dagger$Results adopted from AuViRe~\cite{AuViRe} published paper. 
\end{tablenotes}
\end{threeparttable}
\end{table}

More importantly, the results under tighter thresholds expose the structural flaws of existing paradigms. In particular, as IoU increases from 0.5 to 0.95, the continuous localization approaches experience catastrophic performance degradation. For example, the second-best method, AuViRe~\cite{AuViRe}, drops from AP@0.5 of 98.16\% down to 45.83\% at AP@0.95 on LAV-DF, and similarly drops from 96.96\% down to merely 12.64\% on AV-Deepfake1M. This is mainly because, although localized closely around the forgery segments, the top-ranked proposals fail to deliver accurate boundaries. Meanwhile, the semi-supervised approach relying on cross-modal consistency, MDP~\cite{MDP}, exhibits extreme failure on AV-Deepfake1M, with AP@0.5 of 0.01\%. This is possibly due to that, while semi-supervised consistency analysis identifies trivial full-real or full-fake segments, it lacks the discriminative knowledge to balance for cases where only one modality is forged. On the contrary, our proposed WAFL paradigm favorably retains promising localization precision, with AP@0.95 of 99.31\% on LAV-DF and 97.24\% on AV-Deepfake1M. This practically validates that anchoring proposals to discrete word tokens aligns with the essence of temporal forgeries, bypassing the boundary ambiguity of existing paradigms.

On the other hand, the evaluation on AR@N demonstrates WAFL's capability to prioritize correct forgery proposals with higher confidence scores. Specifically, WAFL achieves nearly perfect recalls at the highest proposal caps, with AR@100 of 99.74\% on LAV-DF and 99.99\% on AV-Deepfake1M. Moreover, such robustness is maintained even under the most strict proposal limits, yielding AR@2 of 99.73\% on LAV-DF and AR@5 of 99.69\%, respectively, outperforming the second-best by 12.00\% and 14.97\%. This confirms that the FFR module and the ACA loss effectively push the actual temporal forgeries to rank the highest confidence. As for existing TFL paradigms, although promising AP@0.5 values demonstrate the right direction towards the forgery neighborhood, the significant decreases from AR@10 to AR@2 in \cref{tab:lavdf} reveal the potential flaw such that the best-matched proposals fail to rank highest. 

\begin{table}[t!]
\caption{Efficiency comparison between different TFL approaches and paradigms for different modules, categorized by `all' and `learnable' parameters. }
\resizebox{\textwidth}{!}{
\begin{threeparttable}
\label{tab:efficiency}
\centering
\begin{tabular}{@{}lcccc>{\cellcolor{gray!20}}c>{\cellcolor{gray!20}}ccc@{}}
\toprule
\multirow{2}{*}{Method} & \multicolumn{2}{c}{Visual Encoder} & \multicolumn{2}{c}{Audio Encoder} & \multicolumn{2}{c}{\cellcolor{gray!20}Localization} & \multicolumn{2}{c}{Full Workflow} \\
\cmidrule{2-9}
 & All & Learnable & All & Learnable & All & Learnable & All & Learnable \\
\midrule
BA-TFD~\cite{BA-TFD} & 2.32 M & 2.32 M & 0.08 M & 0.08 M & \underline{3.10 M} & \underline{3.10 M} & 5.50 M & 5.50 M \\
BA-TFD+~\cite{BA-TFD+} & 47.43 M & 47.43 M & 85.84 M & 85.84 M & 19.62 M & 19.62 M & 152.89 M & 152.89 M \\
UMMAFormer~\cite{UMMAFormer} & 47.04 M & -- & 5.32 M & -- & 49.72 M & 49.72 M & 102.08 M & 49.72 M \\
MDP~\cite{MDP} & 23.51 M & -- & 317.39 M & -- & 13.65 M & 13.65 M & 354.55 M & 13.65 M \\
DiMoDif~\cite{DiMoDif} & 250.38 M & -- & 243.05 M & -- & 6.98 M & 6.98 M & 500.41 M & 6.98 M \\
AuViRe$^\dagger$~\cite{AuViRe} & 101.36 M & -- & 89.86 M & -- & 8.94 M & 8.94 M & 110.38 M & 8.94 M \\
WAFL (ours) & 86.227 M & 0.30 M & 315.439 M & 0.79 M & \textbf{1.45 M} & \textbf{1.45 M} & 403.112 M & 2.54 M \\
\bottomrule
\end{tabular}
\begin{tablenotes}
\item $^\dagger$ Visual and audio encoders of AuViRe share 89.78 M parameters. 
\end{tablenotes}
\end{threeparttable}
}
\end{table}

\subsection{Paradigm Efficiency Analysis}

Existing TFL approaches incur significant parameter overhead when training for forgery boundary regression and continuous frame-level anomaly detection, in addition to the necessary adoption of robust backbone encoders for visual and audio feature extraction. This not only increases the hardware barrier but also makes models vulnerable to overfitting to local biases and noise. By contrast, our proposed WAFL paradigm simplifies the task difficulty of TFL into a precise, binary classification problem. To demonstrate this structural superiority, we presented a comprehensive parameter efficiency analysis in \cref{tab:efficiency}. Specifically, both boundary regression paradigms (BA-TFD~\cite{BA-TFD}, BA-TFD+~\cite{BA-TFD+}, and UMMAFormer~\cite{UMMAFormer}) and frame-level continuous anomaly detection paradigms (MDP~\cite{MDP}, DiMoDif~\cite{DiMoDif}, and AuViRe~\cite{AuViRe}) generally demand a larger number of learnable parameters to finalize the forgery proposals from the visual and audio features. For instance, BA-TFD+~\cite{BA-TFD+} requires updating 152.89 M parameters throughout the entire pipeline, while UMMAFormer~\cite{UMMAFormer} dedicates 49.72 M learnable parameters solely to the localization mechanism. Even recent frameworks like DiMoDif~\cite{DiMoDif} require 6.98 M learnable parameters in their localization heads. 

As for WAFL, it optimizes a remarkably smaller total of only 2.54 M learnable parameters for the entire workflow. Although adopting high-capacity foundation models inevitably increases the total parameter count, the FFR module efficiently bridges the granularity misalignment gap by requiring merely 0.30 M and 0.79 M learnable parameters for the visual and audio encoders, respectively. Furthermore, while adopting state-of-the-art foundation models for feature extraction is a standard convention in the TFL task, WAFL smoothly prevents this from inflating the training cost. Consequently, the forensic discriminative features are processed by only three linear classifier heads for final localization, requiring only 1.45 M learnable parameters. When evaluated alongside the TFL performance in \cref{tab:lavdf} and \cref{tab:avdeepfake1m}, it is evident that, by shifting the paradigm from continuous regression and localization to discrete word-anchored binary classification, WAFL achieves a perfect balance, reaching state-of-the-art localization precision and recall while operating with a promisingly lightweight learnable parameter footprint.

\begin{table}[t!]
\caption{Temporal forgery localization performance (\%) on LAV-DF under cross-dataset setting. \textbf{Best} and \underline{second-best} performance are in \textbf{bold} and \underline{underlined}. }
\resizebox{\textwidth}{!}{
\begin{threeparttable}
\label{tab:avdeepfake1m_lavdf}
\centering
\begin{tabular}{@{}lcccccccc@{}}
\toprule
Method$^\dagger$ & AP@0.5 & AP@0.75 & AP@0.95 & AR@100 & AR@50 & AR@20 & AR@10 & AR@2 \\
\midrule
DiMoDif~\cite{DiMoDif} & \textbf{59.71} & 26.22 & 0.21 & 75.28 & 72.29 & 67.37 & 62.52 & 56.20 \\
AuViRe~\cite{AuViRe} & \underline{48.54} & \underline{39.23} & \underline{0.69} & \underline{77.65} & \underline{76.08} & \underline{73.79} & \underline{70.99} & \underline{58.04} \\
WAFL (ours) & 45.22 & \textbf{44.89} & \textbf{44.89} & \textbf{99.74} & \textbf{99.57} & \textbf{96.31} & \textbf{89.55} & \textbf{69.61} \\
\bottomrule
\end{tabular}
\begin{tablenotes}
\item $^\dagger$Only competitive methods in \cref{tab:lavdf} and \cref{tab:avdeepfake1m} are selected for comparison in cross-dataset evaluation. 
\end{tablenotes}
\end{threeparttable}
}
\end{table}

\subsection{Cross-Dataset Evaluation}

In real-world forensic scenarios, Deepfake detection models are expected to generalize to unseen manipulation techniques and diverse data distributions. However, cross-dataset evaluation has rarely been explored in existing TFL literature. This is primarily because current paradigms still struggle to derive precise boundaries with promising AP@0.95 performance, even in the in-dataset setting. Nevertheless, we argue that cross-dataset evaluation is a critical necessity for the future of TFL. Therefore, to assess generalizability, we adopted the model weights trained on AV-Deepfake1M and evaluated them on LAV-DF. For a fair and meaningful comparison, we only selected the most competitive state-of-the-art methods summarized from \cref{tab:lavdf} and \cref{tab:avdeepfake1m} as baselines, namely DiMoDif~\cite{DiMoDif} and AuViRe~\cite{AuViRe}.

As quantitatively presented in \cref{tab:avdeepfake1m_lavdf}, the cross-dataset performance highlights the critical vulnerability of the continuous anomaly detection paradigm. While DiMoDif and AuViRe achieve moderate AP@0.5 scores of 59.71\% and 48.54\%, respectively, their performance drastically collapses towards near-zero, with merely 0.21\% and 0.69\% at AP@0.95. This indicates that existing paradigms overfit heavily to the local biases and noise of their training datasets. Although roughly guess the forgery neighborhood, they lose the capacity to precisely locate the manipulation boundaries on unseen data. In contrast, our proposed WAFL paradigm demonstrates better boundary robustness across different data distributions. Specifically, by anchoring the localization to discrete word tokens, WAFL avoids the continuous boundary ambiguity that punishes state-of-the-art approaches. Therefore, when requiring precise matching, WAFL achieves an AP@0.95 of 44.89\%, outperforming the second-best, AuViRe~\cite{AuViRe}, by 44.20\%. 

Admittedly, WAFL yields a lower AP@0.5 score of 45.22\% compared to the other baseline methods. This is reasonably expected when transferred to unseen data distributions. In particular, continuous anomaly detectors tend to predict extended, imprecise forgery segments when encountering out-of-distribution noise. Therefore, when the tolerance is high, such as at $\textrm{IoU}=0.5$, the true positive condition is satisfied, while this fails for IoU thresholds that are more strict. Conversely, WAFL enforces strict, discrete lexical boundaries. In other words, when the forensic artifacts of a cross-domain forgery successfully bypass the FFR module during feature extraction, the time ranges of the corresponding word tokens are then completely ignored. This lowers the upper bound of the success rate when the tolerance is high, but ensures that any positive prediction is bounded with the utmost precision, as evidenced by the generally stable AP@IoU across the 0.5, 0.75, and 0.95 thresholds, and the huge advantage at 0.75 and 0.95. 

Furthermore, the AR@N metrics confirm WAFL's strength in identifying and prioritizing true positive proposals, even on unseen data. Specifically, WAFL achieves a highly robust AR@100 of 99.74\% and consistently maintains reliable localization ability, achieving an AR@10 of 89.55\%. Although AR@2 drops to 69.61\%, WAFL still holds a clear advantage, outperforming the second-best, AuViRe, by 11.57\% at the most strict proposal limit. This also verifies that the ACA loss effectively forces the model to assign higher scores to the actual forgery word tokens, whereas the continuous localization paradigms accidentally mess up true positives with a vast number of false positives. 

In summary, by maintaining stability from AP@0.5 to AP@0.95 and demonstrating superior reliability from AR@100 to AR@2, WAFL successfully alleviates the heavy burden of seeking precise boundaries and exhibits state-of-the-art performance under the cross-dataset setting. With the boundaries localized regarding human speech, the true challenge is then revealed: similar to conventional Deepfake detection on images and full videos, improving the generalizability of forensic feature analysis for cross-dataset, cross-manipulation, and cross-domain settings is urgently necessary in the near future of the TFL task.

\subsection{Ablation Study}
\label{sec:ablation_study}

In this section, we conducted ablation sessions to validate the contribution of the core modules in the WAFL paradigm, namely the forensic feature realignment (FFR) module and the artifact-centric asymmetric (ACA) loss. Specifically, we iteratively disabled or replaced the critical components and retrained the models with different configurations for evaluation. The comprehensive quantitative results are presented in \cref{tab:ablation}.

\begin{table}[t!]
\caption{Ablation study of the FFR module and the ACA loss on AV-Deepfake1M. }
\label{tab:ablation}
\centering
\resizebox{\textwidth}{!}{
\begin{tabular}{@{}lcccccccc@{}}
\toprule
Config & AP@0.5 & AP@0.75 & AP@0.95 & AR@100 & AR@50 & AR@20 & AR@10 & AR@5 \\
\midrule
WAFL (optimal) & 97.26 & 97.25 & 97.24 & 99.99 & 99.99 & 99.98 & 99.90 & 99.69 \\
w/o FFR & 43.96 & 43.95 & 43.94 & 99.99 & 99.74 & 96.49 & 89.10 & 79.51 \\
w/ BCE loss & 51.01 & 51.01 & 51.01 & 99.99 & 99.97 & 99.51 & 98.53 & 96.45 \\
w/ Focal loss & 89.94 & 89.92 & 89.92 & 99.99 & 99.97 & 99.70 & 99.13 & 98.17 \\
\bottomrule
\end{tabular}
}
\end{table}

\begin{figure}[tb]
    \centering
    \includegraphics[width=\textwidth]{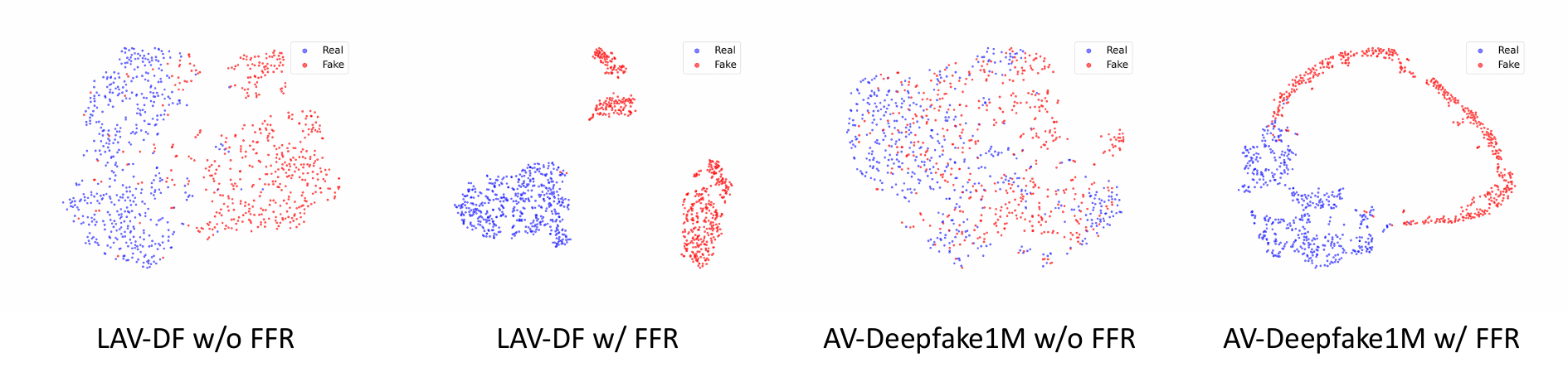}
    \caption{Feature cluster difference caused by the FFR module via t-SNE plots. }
    \label{fig:ffr_cluster}
\end{figure}

\noindent\textbf{Effectiveness of FFR.} In this paper, the FFR module is designed to bridge the gap between low-frequency semantic features and high-frequency forensic artifacts. To verify the necessity, we disabled the module by utilizing only the frozen foundation models for visual and audio feature extraction, while keeping the rest of the model unchanged. It can be observed from \cref{tab:ablation} that (denoted as `w/o FFR'), disabling FFR causes a critical performance damping in localization precision. Specifically, the AP@IoU for all three thresholds drops to approximately 43\%. Furthermore, the most strict recall metric, AR@5, degrades significantly by over 20\%, dropping to 79.51\%. 

To further illustrate the difference, we visualized t-SNE~\cite{t-SNE} scatter plots of the features extracted at the end of the fusion classifier head before dropout and activation, sampling 1,000 random real word tokens and 1,000 random fake word tokens from both LAV-DF and AV-Deepfake1M. As exhibited in \cref{fig:ffr_cluster}, the raw semantic features (`w/o FFR') show abundant overlap between real and fake word tokens, demonstrating that the foundation backbones cannot naturally distinguish forensic artifacts. Nevertheless, by enabling FFR (`w/ FFR'), the features diverge into highly separable real and fake clusters on both datasets. This favorably validates our core hypothesis: while pre-trained foundation models provide rich representations during feature extraction, their semantic space is insensitive to forensic artifacts. Therefore, the FFR module is necessary for adaptively projecting semantic features onto a discriminative forensic manifold.

\noindent\textbf{Effectiveness of ACA Loss.} The ACA loss is particularly designed to mitigate the extreme class imbalance of the forgery word tokens, by dynamically suppressing the overwhelming gradients of real samples. In this section, to demonstrate its superiority, we substituted the optimal ACA loss with standard binary cross-entropy loss (`w/ BCE loss') and standard Focal loss~\cite{FocalLoss} ('w/ Focal loss'). 

As listed in \cref{tab:ablation}, replacing the ACA loss with a standard BCE loss results in a severe performance penalty, yielding an AP@0.5 of only 51.01\%. In particular, although implementing a balanced random sampler to ensure quantity balance in the input batches, the standard BCE loss fails to effectively isolate minority forensic artifacts. Because BCE loss treats all failures equivalently, the extreme imbalance in sample occurrence frequencies between real and fake makes authentic samples dominate the gradient flow.  

Alternatively, leveraging Focal loss partially restores the performance, achieving an AP@0.95 of 89.92\% and an AR@5 of 98.17\%. Nevertheless, while Focal loss successfully down-weights well-classified easy samples, its symmetric nature sets up a critical bottleneck in the current TFL task. In particular, since an identical modulating decay is applied to both real and fake classes, the Focal loss fails to account for the different distributions of occurrence frequencies between real and fake samples. In contrast, the proposed ACA loss is designed with an asymmetric mechanism via $\gamma_+$, $\gamma_-$, and $p_m$ in \cref{eq:aca_loss}, dynamically acknowledging that even subtle forensic artifacts are critical for relentless penalties, while harmless variations of the authentic manifold are relatively forgivable and should have their gradients truncated. To conclude, as illustrated in \cref{tab:ablation}, the proposed optimal WAFL configuration consistently achieves the best performance with an AP@0.95 of 97.24\% and an AR@5 of 99.69\%, proving the necessity of the proposed ACA loss. 

\section{Conclusion}
\label{sec:conclusion}

In this paper, we introduce word-anchored temporal forgery localization (WAFL), a novel paradigm that shifts the conventional approaches of the temporal forgery localization task from boundary regression and continuous localization to discrete word-level binary classification. To address the granularity misalignment between the semantic feature space of foundation encoder models and the discriminative forensic manifold, we propose the forensic feature realignment (FFR) module. Furthermore, to mitigate extreme class imbalance between real and fake, we design the artifact-centric asymmetric (ACA) loss, which dynamically suppresses overwhelming authentic gradients while asymmetrically prioritizing subtle forensic artifacts. Extensive experiments demonstrate that WAFL achieves state-of-the-art temporal forgery localization performance on the LAV-DF and AV-Deepfake1M datasets even across stringent AP@IoU and AR@N metrics. By alleviating the computational burden of continuous boundary regression through lexical anchoring, WAFL redefines the baseline for precise and efficient temporal forgery localization on multimodal audio-visual Deepfake videos. 

Moreover, this novel paradigm allows future research to pivot toward the critical challenge of forensic feature generalizability. While our framework significantly outperforms existing state-of-the-art methods, extracting domain-agnostic forensic artifacts remains an open challenge for temporal forgery localization. Additionally, we acknowledge the potential limitation that WAFL relies on off-the-shelf speech-to-text alignment tools to derive reliable initial lexical boundaries. However, rather than a vulnerability, we view this as a scalable advantage. Motivated by rapid and continuous advancements in multimodal modeling, these auxiliary tools will only become more precise. Therefore, the WAFL paradigm paves a robust and promising direction for the future of defending against temporal forgeries in audio-visual Deepfake videos. 

% \begin{table}[t!]
% \caption{Method efficiency comparison regarding learnable parameters, FLOPs, and Throughput. }
% \label{tab:efficiency_trainable}
% \centering
% % \resizebox{\textwidth}{!}{
% \begin{tabular}{@{}lccccc@{}}
% \toprule
% \multirow{2}{*}{Method} & \multicolumn{2}{c}{Localization} & \multicolumn{3}{c}{Full Workflow} \\
% \cmidrule{2-6}
% & Params & FLOPs & Params & FLOPs & Throughput \\
% \midrule
% BA-TFD~\cite{BA-TFD} & 3.104 M & 2002.060 G & 5.502 M & 2005.800 G & 426.96 s \\
% BA-TFD+~\cite{BA-TFD+} & 19.621 M & 403.106 G & 152.898 M & 775.576 G & 329.88 s \\
% UMMAFormer~\cite{UMMAFormer} & 49.723 M & 46.929 G & 49.723 M & 28.288 G & \\
% MDP~\cite{MDP} & 13.647 M & 4.980 G & 13.647 M & 188.318 G & \\
% DiMoDif~\cite{DiMoDif} & 6.978 M & 3.194 G & 6.978 M & 193.134 G & \\
% AuViRe~\cite{AuViRe} & 8.935 M & 37.060 G & 8.935 M & & \\
% WAFL (ours) & 1.446 M & 0.001 G & 2.527 M & 292.655 G & 29.82 s \\
% \bottomrule
% \end{tabular}
% % }
% \end{table}

% \clearpage  % TODO FINAL: This \clearpage needs to be removed from both review and camera-ready versions.

% \section*{Acknowledgements}
% Please insert your acknowledgments here.

\bibliographystyle{splncs04}
\bibliography{main}
\end{document}